\documentclass[sigconf, nonacm]{acmart}

\makeatother
      
\usepackage{fancyhdr}
\usepackage{multirow} 
\usepackage{amssymb}
\usepackage[ruled]{algorithm2e}
\usepackage{caption}
\usepackage{graphicx}
\usepackage{float} 
\usepackage{subfigure}
\usepackage{wrapfig}
\usepackage{hyperref}

\begin{document}



\title{Efficient Bilateral Cross-Modality Cluster Matching for Unsupervised Visible-Infrared Person ReID}

\author{De Cheng$^{1}$\footnotemark[1], Lingfeng He$^{1}$\footnotemark[1], Nannan Wang$^{1}$\footnotemark[2], \\
Shizhou Zhang$^{2}$,
Zhen Wang$^{3}$, Xinbo Gao$^{4}$,\\
$^{1}$ Xidian University,
$^{2}$ Northwestern Polytechnical University,\\
$^{3}$ Zhejiang Lab,
$^{4}$ Chongqing University of Posts and Telecommunications\\
}
\renewcommand{\shortauthors}{}








\begin{abstract}

Unsupervised visible-infrared person re-identification (USL-VI-ReID) aims to match pedestrian images of the same identity from different modalities without annotations. Existing works mainly focus on alleviating the modality gap by aligning instance-level features of the unlabeled samples. However, the relationships between cross-modality clusters are not well explored. To this end, we propose a novel bilateral cluster matching-based learning framework to reduce the modality gap by matching cross-modality clusters. Specifically, we design a Many-to-many Bilateral Cross-Modality Cluster Matching (MBCCM) algorithm through optimizing the maximum matching problem in a bipartite graph. Then, the matched pairwise clusters utilize shared visible and infrared pseudo-labels during the model training. Under such a supervisory signal, a Modality-Specific and Modality-Agnostic (MSMA) contrastive learning framework is proposed to align features jointly at a cluster-level. Meanwhile, the cross-modality Consistency Constraint (CC) is proposed to explicitly reduce the large modality discrepancy. Extensive experiments on the public SYSU-MM01 and RegDB datasets demonstrate the effectiveness of the proposed method, surpassing state-of-the-art approaches by a large margin of 8.76\% mAP on average. 

\end{abstract}

\keywords{USL-VI-ReID, Cluster-Level, Bipartite Graph, Modality Discrepancy}

\maketitle

\section{Introduction}

\begin{figure}
\centering
\includegraphics[width=0.46\textwidth]{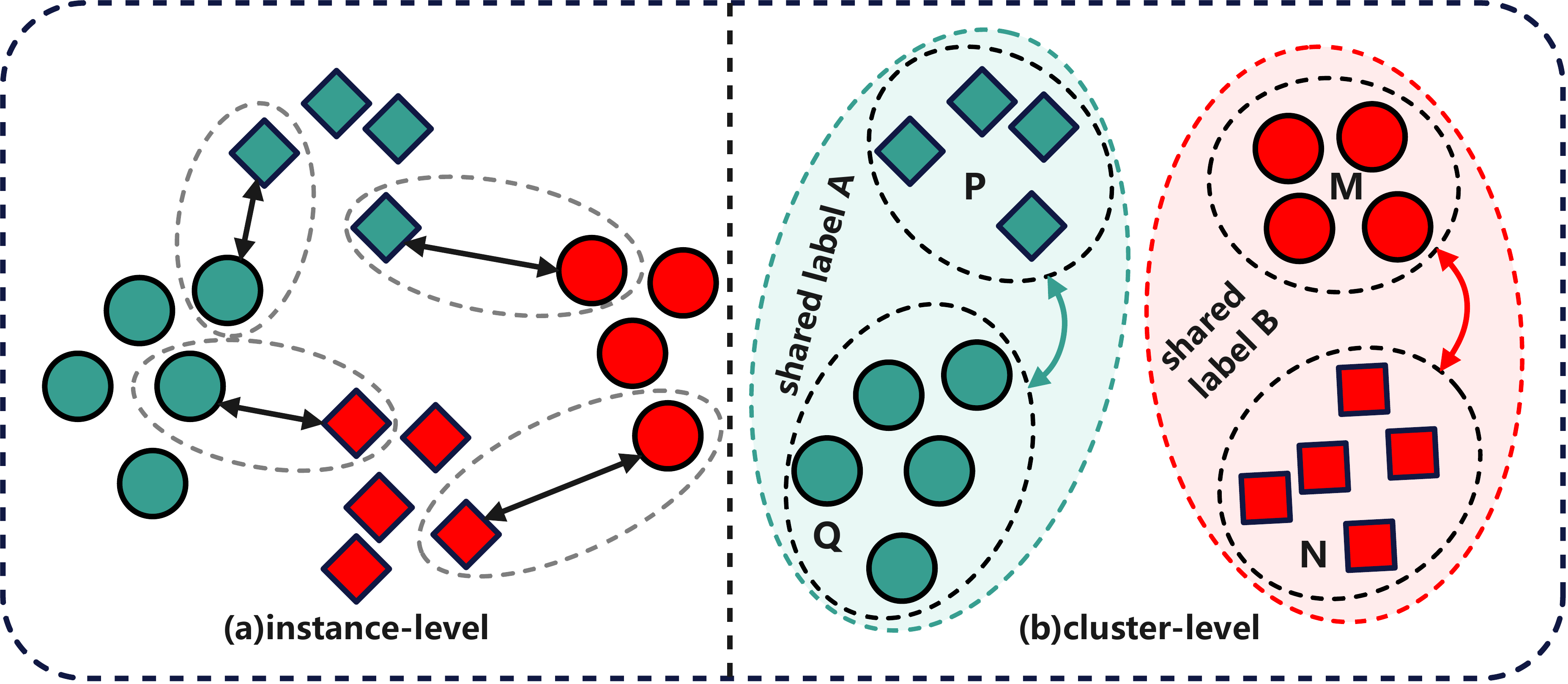}
\caption{The information exchange between modalities is a vital factor that affects the performance of VI-ReID. Existing methods always utilize the information between pairwise similar instances at an \textbf{(a)} instance-level, which cannot holistically build relationships between cross-modality classes. To address this issue, we propose a \textbf{(b)} cluster-level matching and generate shared-label cross-modality clusters that provide supervision for network training.} 
\label{cluster-instance}
\end{figure}

Visible-infrared person re-identification (VI-ReID) \cite{wu2017rgb, bai2017scalable, 2022OLTA, ye2018hierarchical, Hou_2019_CVPR, 2021Farewell} refers to the task of retrieving person images from a visible/infrared camera corresponding to a given query from another modality. This task is attracting increasing attention due to its widespread application in intelligent surveillance systems. Compared to the widely studied single-modality person ReID~\cite{sreid1, sreid2, sreid3, sreid4, sreid5, sun2018beyond, li2018harmonious}, VI-ReID is a more challenging problem on account of the large modality gap between visible and infrared images.

Existing methods~\cite{2021channel, 2022MAUM, wu2021discover, ye2020dynamic, wang2019rgb, yang2022learning, park2021learning} generally address the large modality gap in VI-ReID task from two perspectives: (1) reducing the modality gap at the image level by diverse augmentations and GAN-based image generation \cite{2021channel, wang2019rgb, 2023LTG}; (2) designing novel loss functions or network structures to obtain modality-invariant features \cite{wu2021discover, 2022CIFT}. 
Although these works have achieved remarkable performance based on datasets with shared annotations across modalities, 
it is always very expensive and time-consuming to annotate the cross-modality identities in real-world scenarios.
For these reasons, this work focuses on VI-ReID under a pure unsupervised setting (USL-VI-ReID), which is more data-friendly than conventional supervised VI-ReID.

The key to USL-VI-ReID task is to establish data associations between modalities without identity annotations.
Some works perform cross-modality information interaction through neighbor instances. H2H \cite{2021H2H} designed an ISML loss to narrow the distances among cross-modality neighbor instances, while ADCA \cite{2022ADCA} selected pairwise instances with high similarities to perform cross-modality instance associations. Unlike the above instance-wise methods, OTLA \cite{2022OLTA} built the relationship between infrared instances and visible identities through an optimal-transport assignment. These methods achieve modality fusion to a certain extent. However, such methods  only consider instance-level relationships while neglecting interrelation at the cluster-level, which cannot provide a stable and systematic supervisory signal to the learning framework. Due to the remarkable development of unsupervised single-modality ReID (e.g., Cluster-Contrast \cite{2021ClusterContrast}, which achieved 82\% mAP and 92\% Rank-1 on Market-1501), high-quality clusters can be obtained under visible and infrared modalities, respectively. Thus, a cluster-level matching scheme as shown in Figure \ref{cluster-instance}, which builds relationships between these high-quality single-modality clusters, has the potential to achieve excellent performance for USL-VI-ReID task.

Based on the aforementioned motivations, we propose a bilateral cluster matching based learning framework, a novel unsupervised VI-ReID framework that provides cross-modality shared pseudo-labels through an effective cluster-level matching paradigm. 
We assume that as the training process goes on, the ability of the model to extract modality-invariant features is continuously reinforced, and the cross-modality clusters with the same identity get closer 
to each other in the embedding space.
Based on this assumption, we construct a weighted bipartite graph between pairwise cross-modality clusters. Then, we propose two maximum matching problems by regarding visible and infrared clusters as queries, respectively. Such bilateral matching ensures each cluster gets matched, but this one-to-one matching paradigm is not robust enough to overcome matching noise and intra-class variations. 
To address this problem, we propose an efficient Many-to-many Bilateral Cross-modality Cluster Matching (MBCCM) mechanism. During the training stage, pairwise instances with shared visible and infrared labels sampled from the matching matrix are sent to the encoder, and the shared labels maintain a stable supervisory signal in one training epoch.

The proposed MBCCM mechanism transforms the USL-ReID task into a supervised problem by providing a supervisory signal to the training stage. However, the conventional supervised task cannot be directly applied due to the different cluster numbers for different modalities. To address this, we propose a Modality-Specific and Modality-Agnostic (MSMA) learning framework to perform contrastive learning under such an unsupervised pseudo-label-based method. Specifically, we construct two memory banks specific to each modality and two modality-agnostic memory banks to overcome the inconsistency of the cluster number between different modalities. Additionally, 
the cross-modality consistency constraint aims to keep the consistency of the predictions from corresponding memory banks 
and is proposed to explicitly reduce the large modality discrepancy. 

In summary, our main contributions are as follows:

\begin{itemize}
\item We propose a novel many-to-many bilateral cross-modality cluster matching  algorithm (MBCCM), through optimizing the maximum matching problem in a bipartite graph, to build relationships between cross-modality clusters for USL-VI-ReID task.
\item We design a Modality-specific and Modality-agnostic (MSMA) contrastive learning framework followed by the cross-modality consistency constraint, to effectively reduce modality discrepancy in USL-VI-ReID. 
\item Extensive experiments on the public SYSU-MM01 and RegDB datasets demonstrate the effectiveness of the proposed method, which surpasses state-of-the-art method by a large margin of +5.43\%/+13.82\% mAP on the SYSU-MM01 and RegDB benchmarks, respectively.
\end{itemize}

\section{Related Work}

\begin{figure*}
  \centering
  \includegraphics[width=17.8cm]{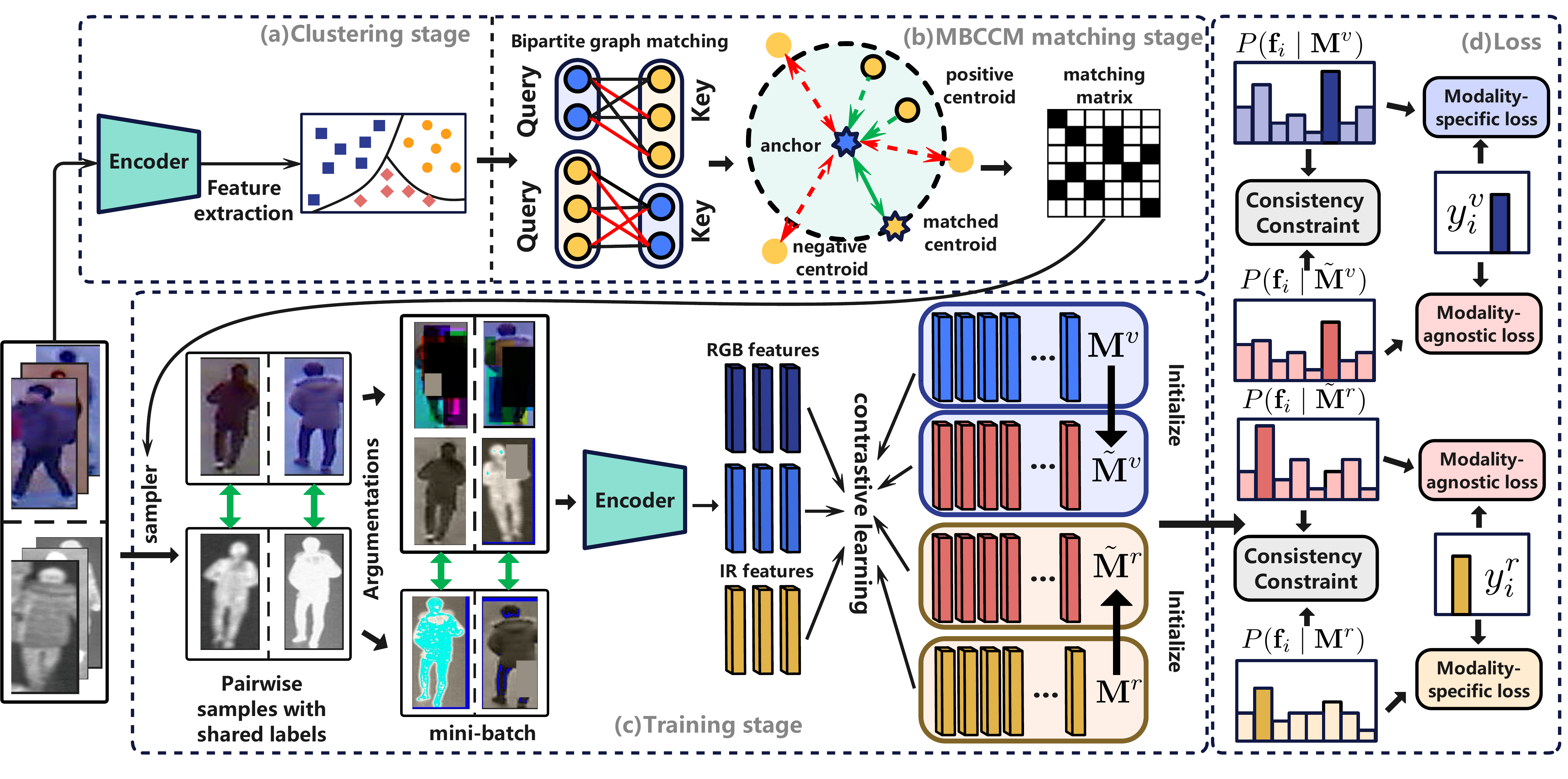}\\
  \caption{The overall framework of our method. Our method consists of a clustering stage, a matching stage, and a training stage. In the clustering stage (a), we assign pseudo labels to samples from each modality. In the matching stage (b), we utilize Many-to-many Bilateral Cross-Modality Centroid Matching (MBCCM) to perform a cluster-level match between cross-modality clusters. In the training stage (c)/(d), modality-specific and modality-agnostic (MSMA) memory banks jointly construct a contrastive learning framework, and the Consistency Constraint (CC) module further reduces the modality gap.}\label{framework}
\end{figure*}

\subsection{Supervised Cross-modality person ReID}

Visible-infrared person re-identification (VI-ReID) aims to retrieve person images with the same identity from visible and infrared cameras, which is a challenging problem due to the significant modality gap between visible and infrared images \cite{bai2017scalable, ye2018hierarchical, Hou_2019_CVPR, Hao_2021_ICCV, 2021Farewell, 2020Bi, ye2020dynamic, 2022ADCA, wu2021discover, 2022OLTA}. Wu \textit{et al.} \cite{wu2017rgb} first introduces the problem of cross-modality person ReID and proposes a zero-padding network. Some other works tried different image augmentation strategies to reduce the modality discrepancy~\cite{2023LTG, 2021channel, qian2023visibleinfrared, josi2022multimodal}. CA \cite{2021channel} generates an intermediate modality by a random channel argument method. LTG \cite{2023LTG} introduces a linear transformation generator based on the Lambertian model. Another branch of works designs novel loss functions or network structures to extract modality-invariant features. MAUM~\cite{2022MAUM} proposes a memory-based augmentation to pull close the features in the counterpart modality. CIFT~\cite{2022CIFT} utilizes the counterfactual intervention to eliminate the modality gap.

The above-mentioned works use various methods to narrow the modality gap, which has been proven useful for VI-ReID. 
However, these methods mainly focus on the supervised VI-ReID problem, which requires a large amount of image annotations, making them too expensive for practical applications. In this work, we investigate USL-VI-ReID based on the aforementioned methods, aiming to learn modality-invariant features without annotations.

\subsection{Unsupervised single-modality person ReID}

Unsupervised single-modality ReID (USL-ReID) aims to learn robust feature representations for unlabeled person images within a single modality \cite{uslreid1, uslreid2, uslreid3, uslreid4, uslreid5, 2021ClusterContrast, 2021ICE, 2022PPLR}. State-of-the-art unsupervised ReID methods typically consist of two stages: 1) generating pseudo labels through clustering; and 2) training a model with the pseudo labels under a supervised paradigm. To refine the noisy labels \cite{2020MMT, 2022PPLR} generated from clustering, MMT \cite{2020MMT} employs auxiliary teacher networks to generate soft labels for better supervision. PPLR \cite{2022PPLR} refines the pseudo labels by using the relationship between global features and part features. Memory-based methods~\cite{2020spcl, 2021ClusterContrast, 2022ISE, 2021ICE} achieve promising performances on unsupervised ReID. SPCL \cite{2020spcl} gradually generates more robust clusters through a self-paced learning framework with a hybrid memory bank. To overcome the cluster imbalance, Cluster-Contrast \cite{2021ClusterContrast} constructs a cluster-based memory bank that uses a unique representation to describe each cluster. 

On top of the above two-stage approach and the concept of memory banks, we construct unsupervised cluster-based modality-specific and modality-agnostic memory banks to enhance contrastive learning between the two modalities.

\subsection{Unsupervised Cross-modality person ReID}

Unsupervised Cross-modality person ReID expands on the mission of USL-ReID to include visible and infrared modalities, with the goal of extracting modality-invariant features under noisy pseudo-labels. H2H \cite{2021H2H} introduces a framework consisting of a homogeneous stage and a heterogeneous stage, while OTLA \cite{2022OLTA} treats the assignment between infrared instances and visible labels as an optimal transport problem. 
ADCA \cite{2022ADCA} proposes a framework with two modality-specific memory banks and a process of cross-modality memory aggregation, which connects memory banks under different modalities. 

In contrast to above-mentioned works, our approach begins with the unsupervised clustering results and then proposes to explore the cluster-level relationships between cross-modality clusters.

\section{Proposed method}

\subsection{Overview}\label{baseline}


Let $\mathcal{X} = \left\{\mathcal{V}, \mathcal{R}\right\}$ denote an unlabeled cross-modality ReID dataset, where $\mathcal{V} = \left\{\mathbf{x}^v_i\right\}^{N_v}_{i=1}$ and $\mathcal{R} = \left\{\mathbf{x}^r_i\right\}^{N_r}_{i=1}$ denote $N_v$ visible images and $N_r$ infrared images from two modalities, respectively. In the USL-VI-ReID task, our goal is to train a deep neural network $f_{\theta}(\cdot)$ to project an image $\mathbf{x}_i$ from the dataset $\mathcal{X}$ into an embedding space $\mathcal{F}_{\theta}$ and obtain a d-dimensional modality-invariant representation $\mathbf{f}_i = f_{\theta}(\mathbf{x}_i) \in \mathbb{R}^{d}$. 

Following recent clustering-based USL-ReID methods~\cite{2022PPLR, 2022ISE, 2021ClusterContrast, 2020spcl, 2021ICE, dai2021idm, fu2019self, 2022ADCA}, we first utilize DBSCAN \cite{1996DBSCAN} to obtain pseudo labels $\mathcal{Y_V} = \left\{{y}^v_i\right\}^{N_v}_{i=1}$ and $\mathcal{Y_R} = \left\{{y}^r_i\right\}^{N_r}_{i=1}$ for unlabeled samples from two modalities.
Following the memory-bank-based framework ADCA~\cite{2022ADCA}, we use a two-stream encoder with modality-specific shallow layers and modality-shared deep layers to extract features as our baseline. 
Specifically, we generate an intermediate-modality set $\hat{\mathcal{V}} = \left\{\mathbf{x}^{\hat{v}}_i\right\}^{N_v}_{i=1}$ based on the visible images following techniques in CA \cite{2021channel}, then $\left\{\mathbf{x}^v_{i}, \mathbf{x}^{\hat{v}}_{i}, \mathbf{x}^r_{i}\right\}$ constitutes a mini-batch and is sent to the encoder. Meanwhile, we initialize two modality-specific memory banks $\mathbf{M}^v$ and $\mathbf{M}^r$ by the cluster centroids $\mathbf{M}^v = \left\{\mathbf{c}^v_i\right\}_{i=1}^{K_v}$ and $\mathbf{M}^r = \left\{\mathbf{c}^r_i\right\}_{i=1}^{K_r}$ respectively, to store the representations of clusters from each modality and update them with a momentum strategy~\cite{2021ClusterContrast, 2021ICE, 2020spcl, 2022ISE}, where $K_v$ and $K_r$ are the numbers of grouped clusters in these two modalities.
This training process enables the encoder to extract expressive features for each modality and generate high-quality pseudo labels. 

The overall framework is illustrated in Figure \ref{framework}. To integrate information from both modalities, we propose a Many-to-many Bilateral Cross-modality Centroid Matching (MBCCM) module (upper right in Figure \ref{framework}) to match clusters between modalities and provide shared pseudo visible and infrared labels for heterogeneous instances.
The MBCCM module has strong robustness to matching noise by establishing complex many-to-many cross-modality clustering matching relationships.
Spontaneously, two modality-agnostic memory banks are designed correspondingly based on the two modality-specific memory banks. Then, a Modality-Specific and Modality-Agnostic (MSMA) learning framework is proposed to perform cross-modality contrastive learning, followed by a cross-modality Consistency Constraint (CC) module. 
The CC module constrains the consistency of predictions from corresponding modality-specific and modality-agnostic memory banks to strengthen the invariance of the features extracted from different modalities.

\subsection{Many-to-Many Bilateral Cross-modality Cluster Matching (MBCCM)}\label{BCCM}

The interaction of information between cross-modalities is a crucial factor that affects the performance of the VI-ReID model. Matching clusters between the two modalities can help establish relationships between cross-modality clusters, providing a reasonable basis for subsequent information transmission. The basic idea for cluster matching is to select the cross-modality cluster with the highest similarity. However, this operation has two disadvantages: 1) Infrared images contain less information than visible images, which may be limited to a small subspace in the high-dimensional embedding space. This results in an unbalanced match, where most of the infrared clusters can only be matched to a small number of specific visible clusters; 2) This method cannot guarantee that each cluster is matched, which may lead to the exclusion of some clusters during training. These underlying problems lead to inadequate information utilization under direct matching.

\begin{figure}
\centering
\includegraphics[width=0.47\textwidth]{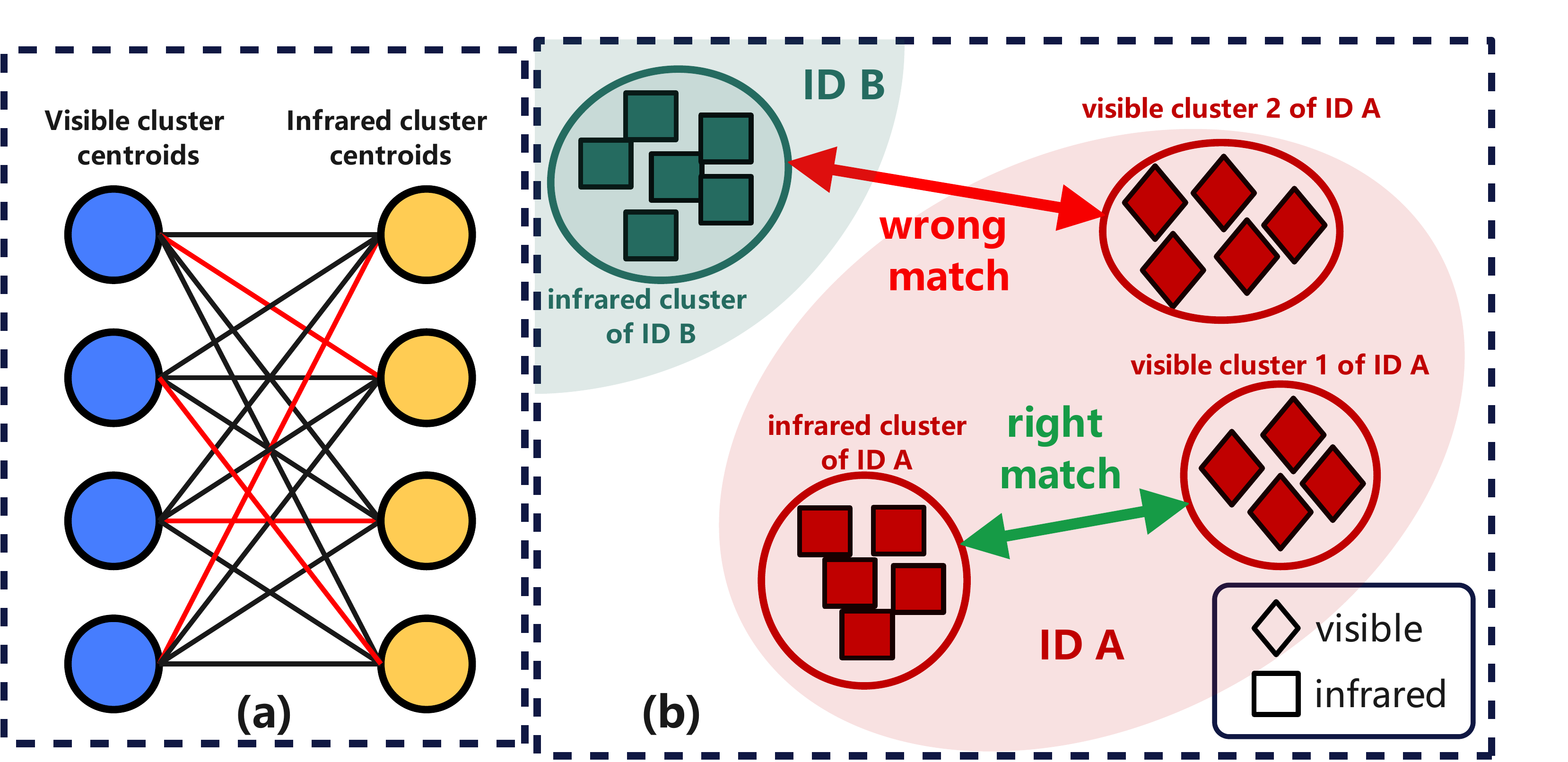}
\caption{(a) A sketch map of the maximum matching problem in a weighted bipartite graph. (b) An example of a wrong match is shown, where two visible clusters with the same ID are matched to different infrared clusters under the one-to-one matching paradigm.} 
\label{match}
\end{figure}

To address these issues, we propose a Bilateral Cross-Modality Cluster Matching (BCCM) mechanism. We formulate the cluster matching problem into a maximum matching problem with a weighted bipartite graph. For visible clusters, we aim to match each of them to an infrared cluster while ensuring the minimum cost, as shown in Figure \ref{match}(a). This can be formulated as a graph optimization problem as follows:
\begin{equation}\label{optimize visible}
\begin{aligned}
&\min_{\mathbf{Q^v}} \ \sum_{i}\sum_{j}\mathbf{P}_{ij} \cdot \mathbf{Q}^v_{ij},\\
&\min_{\mathbf{Q^r}} \ \sum_{i}\sum_{j}\mathbf{P}_{ij} \cdot \mathbf{Q}^r_{ij},\\
s.t.\quad  & \left\{\begin{array}{lc}
\vspace{1ex}
\sum_{j=1}^{K_r}\mathbf{Q}^v_{ij} = 1, & i = 1,2 \cdots, K_v, \\
\vspace{1ex}
\sum_{i=1}^{K_v}\mathbf{Q}^r_{ij} = 1, & j = 1,2 \cdots, K_r, \\
\mathbf{Q}^v_{ij}, \mathbf{Q}^r_{ij} \in \left \{0,1\right \}, \\
\end{array}\right.
\end{aligned}
\end{equation}
where $\mathbf{P} \in \mathbb{R}^{K_{v}\times K_{r}}$ represents the cost matrix, and the element $\mathbf{P}_{ij}$ in the matrix represents the cost needed to match the visible centroid $\mathbf{c}^{v}_i$ to the infrared centroid $\mathbf{c}^{r}_j$. 
We calculate the cost by using the Euclidean distance between the cluster centroids from the two modalities, which is formulated as follows:
\begin{equation}\label{distance}
\mathbf{P}_{ij} = \sqrt
{\Vert \mathbf{c}^{v}_i \Vert^{2}_{2} +
\Vert \mathbf{c}^{r}_j \Vert^{2}_{2} -
2(\mathbf{c}^{v}_i)^{T}(\mathbf{c}^{r}_i)}.
\end{equation}
$\mathbf{Q}^v \in \mathbb{R}^{K_{v}\times K_{r}}$ represents the matching relationship between the cluster centers in the two modalities, where $\mathbf{Q}^v_{ij} = 1$ represents that $\mathbf{c}^{v}_i$ and $\mathbf{c}^{r}_j$ are matched, and thus the instances in these two clusters can use a shared pseudo-label during network training. Similarly, $\mathbf{Q}^r \in \mathbb{R}^{K_{v}\times K_{r}}$ ensures that each infrared cluster can be matched to one visible cluster. The two problems mentioned in Eq. \ref{optimize visible} can be solved using the Kuhn-Munkres (K-M) algorithm (Kuhn 1955; Munkres 1957). Notably, the visible clusters and infrared clusters are regarded as the query respectively, to make a one-to-one matching, so called bilateral matching. 


The BCCM provides a reasonable and relatively uniform matching between cross-modality clusters from a global perspective. Nevertheless, two intuitive drawbacks are revealed: 1) The matched cluster is very likely not an ideal local match; it is just a result of a compromise on the minimum global cost; 2) Due to the intra-class variation, images of the same ID may be clustered into multiple different clusters. In the aforementioned one-to-one matching, clusters from the same ID will be wrongly matched to multiple cross-modality clusters from other IDs, as illustrated in Figure \ref{match}(b). 

To solve these issues, a many-to-many BCCM (MBCCM) strategy is proposed based on the aforementioned one-to-one matching. For each cluster, cross-modality clusters that have smaller distances compared to the matched cluster will be denoted as matched clusters. Indexes of these extended clusters can be formulated as:
\begin{equation}\label{extend matching}
i^m_v = \mathop{\arg}\limits_{i}
[d(\mathbf{c}^{v}_{k_v} , \mathbf{c}^r_{i}) \leq d(\mathbf{c}^{v}_{k_v} , \mathbf{c}^r_{i^{\ast}})]
\end{equation}
where $\mathbf{c}^r_{i^{\ast}}$ denotes the infrared cluster matched by the $k_v$-th visible cluster. And $d(u,v)$ denotes the Euclidean distance between vector $u$ and $v$. 
The precise matching relationships between samples are preserved while more relationships between cross-modality clusters are established, which alleviates the side effects caused by incorrect matching to some extent.

Now we obtain the extended many-to-many matching matrix $\mathbf{Q}^{v}$ and $\mathbf{Q}^{r}$, which ensures that each cluster has at least one matching cluster. Additionally, this match should be mutual. Using $\mathbf{c}^{v}_i$ as the key to match gallery $\mathbf{c}^{r}_j$, it should also be reasonable to match $\mathbf{c}^{v}_j$ with $\mathbf{c}^{r}_i$ as the key. Thus, the final matching matrix is obtained by computing the logical OR between $\mathbf{Q}^{v}$ and $\mathbf{Q}^{r}$, $\mathbf{Q}^{v} \vee \mathbf{Q}^{r}$. Algorithm \ref{mbccm_algori} represents the complete process of bilateral cross-modality cluster matching. During training, we sample pairwise matched clusters ${a,b}$ when $Q_{ab}=1$. Then a visible instance $\mathbf{f}^v_i$ from the a-th cluster and an infrared instance $\mathbf{f}^r_j$ from the b-th cluster share a group of pseudo label $\{y^v_i, y^r_j\}$ to perform contrastive learning. 
\begin{algorithm}
    \caption{A MBCCM cluster-level matching}
    \label{mbccm_algori}
    \LinesNumbered 
	\KwIn{
    \\ visible cluster centroids $\mathbf{M}^{v} \in \mathbb{R}^{K_{v} \times d}$
    \\ infrared cluster centroids $\mathbf{M}^{r} \in \mathbb{R}^{K_{r} \times d}$
    }
	\KwOut{bilateral matching matrix $\mathbf{Q} \in \mathbb{R}^{K_{v}\times K_{r}}$}
    Compute cost matrix $\mathbf{P} \in \mathbb{R}^{K_{v}\times K_{r}}$
    according to Eq. \ref{distance}; \\
    Solve the unidirectional matching problem and get $\mathbf{Q}^{v}$ , $\mathbf{Q}^{r}$ described in Eq. \ref{optimize visible}; \\
    \For{$k_v$ \textbf{in} [1, $K_{v}$]}{
        Get the matched infrared cluster $\mathbf{c}^r_{i^{\ast}}$ from $\mathbf{Q}^v$ where $i^{\ast}=\mathop{\arg}\limits_i(\mathbf{Q}^{v}[k_v,i^{\ast}]==1)$;\\
        Acquire extended matched indexes $i^m_v$ according to Eq. \ref{extend matching};\\
        mark the matching matrix $\mathbf{Q}^{v}[k_v,i^m_v] = 1$;} 
    \For{$k_r$ \textbf{in} [1, $K_{r}$]}{
        Get the matched visible cluster $\mathbf{c}^v_{i^{\ast}}$ from $\mathbf{Q}^r$ where $i^{\ast}=\mathop{\arg}\limits_i(\mathbf{Q}^{r}[i^{\ast},k_r]==1)$;\\
        Acquire extended matched indexes $i^m_r$ according to Eq. \ref{extend matching};\\
        mark the matching matrix $\mathbf{Q}^{r}[i^m_r, k_r] = 1$;}
    compute $\mathbf{Q} = \mathbf{Q}^{v} \vee \mathbf{Q}^{r}$ 
\end{algorithm}

\subsection{Modality-specific and Modality-agnostic (MSMA) Contrastive Learning}\label{MSMA}

Thanks to the matching matrix generated by the MBCCM module, the relationship between cross-modality clusters can be better utilized in contrastive learning. Two modality-agnostic memory banks are denoted as $\tilde{\mathbf{M}}^{v} = \left\{\tilde{\mathbf{c}}^v_i\right\}_{i=1}^{K_v}$ and 
$\tilde{\mathbf{M}}^{r} = \left\{\tilde{\mathbf{c}}^r_i\right\}_{i=1}^{K_r}$. 
The visible-based memory bank $\tilde{\mathbf{M}}^{v} \in \mathbb{R}^{K_{v} \times d}$ is initialized with the centroids of visible clusters ${\{\mathbf{c}}^v_i\}_{i=1}^{K_v}$, thus having the same scale as $\mathbf{M}^v$, 
while the infrared-based memory bank is initialized with the infrared cluster centroids ${\{\mathbf{c}}^r_i\}_{i=1}^{K_r}$. 
Thus, a Modality-specific and Modality-agnostic contrastive learning framework is constructed jointly using these four memory banks. For a visible instance $\mathbf{f}^{v}_i$, 
the modality-specific loss is formulated as: 
\begin{equation}\label{Contrastive1}
\mathcal{L}_{v\rightarrow \mathbf{M}^v}=-\log \frac{\exp \left(sim(\mathbf{f}^{v}_i , \mathbf{c}^v_{+})/ \tau\right)}{\sum_{k=1}^{K_{v}} \exp \left(sim(\mathbf{f}^{v}_i, \mathbf{c}^v_k) / \tau\right)},
\end{equation}
where $\mathbf{c}^v_{+}$ indicates the positive prototype corresponding to label $\mathbf{y}^{v}_i$ in memory $\mathbf{M}^v$, and $\tau$ is a temperature hyper-parameter. $sim(u,v) = u^Tv/\Vert u \Vert \Vert v \Vert$ denotes the cosine similarity between vector $u$ and $v$.
The contrastive losses between the instance and two modality-agnostic memories can be analogously formulated.
Following the momentum updating stratege\cite{2021ClusterContrast, 2022ADCA, 2022ISE}, the instances are applied to update one modality-specific memory corresponding to its modality and two modality-agnostic memories:
\begin{equation}\label{update}
    \begin{aligned}
         \mathbf{c}^v_{y_i}\leftarrow \mu \mathbf{c}^v_{y_i} + (1 - \mu) \mathbf{f}^{v}_{y_i},
    \end{aligned}
\end{equation}
where $\mu$ is the momentum updating factor, $\mathbf{c}^v_{y_i}$ is the $y_i$-th prototype in the memory bank, $\mathbf{f}^{v}_{y_i}$ is the input visible instance feature which belongs to the $y_i$-th cluster in current mini-batch.
Note that modality-agnostic memories are initialized the same as their corresponding modality-specific memories.
However, modality-specific memories are updated only with instances from the corresponding modality, while modality-agnostic memories are updated with instances from both modalities.

For instances in a mini-batch denoted as $\{\mathbf{f}_i^{v}, \mathbf{f}_i^{\hat{v}}, \mathbf{f}_i^{r}\}$, the total modality-specific loss $\mathcal{L}_{ms}$ and modality-agnostic loss $\mathcal{L}_{ma}$ are formulated as Eq. \ref{Lms} and Eq. \ref{Lma}:
\begin{equation}\label{Lms}
\mathcal{L}_{ms} =
\mathcal{L}_{v\rightarrow \mathbf{M}^v} + 
\mathcal{L}_{\hat{v}\rightarrow \mathbf{M}^v} + 
\mathcal{L}_{r\rightarrow \mathbf{M}^r},
\end{equation}
\begin{equation}\label{Lma}
\begin{aligned}
\mathcal{L}_{ma} =
\mathcal{L}_{v\rightarrow \tilde{\mathbf{M}}^v} +
\mathcal{L}_{\hat{v}\rightarrow \tilde{\mathbf{M}}^v} +
\mathcal{L}_{r\rightarrow \tilde{\mathbf{M}}^v}
+
\mathcal{L}_{v\rightarrow \tilde{\mathbf{M}}^r} +
\mathcal{L}_{\hat{v}\rightarrow \tilde{\mathbf{M}}^r} +
\mathcal{L}_{r\rightarrow \tilde{\mathbf{M}}^r},
\end{aligned}
\end{equation}
where the subscript $p \rightarrow \mathbf{M}$ represents in $\mathcal{L}_{p\rightarrow \mathbf{M}}$ we perform contrastive learning between feature $\mathbf{f}^{p}_i$ and memory $\mathbf{M}$.

\subsection{Cross-modality Consistency Constraint (CC)}\label{CCP}

An encoder that performs well must have a strong ability to extract modality-invariant features. Based on the initial information exchange in MSMA learning, we expect the encoder to extract cross-modality features with higher similarity. A simple conjecture is that as the encoder parameters are updated, the clustering stage will provide relatively robust shared pseudo labels for the training stage. Therefore prototypes from the corresponding modality-specific and modality-agnostic memories with the same pseudo label can be regarded as two representations of the same ID. To facilitate cross-modality invariance, the representations of one ID from different memory banks are expected to be consistent. To this end, for each instance, we constrain the consistency of its predictions from a modality-specific and the corresponding modality-agnostic memory banks according to the Kullback-Leibler divergence. For $\mathbf{f}^{v}_i$, the consistency constraint between $\mathbf{M}^v$ and $\tilde{\mathbf{M}}^v$ can be formulated as:
\begin{equation}\label{KLDivergencevva}
\begin{aligned}
\mathcal{L}^{v}_{\mathbf{M}^v\leftrightarrow \tilde{\mathbf{M}}^v}= 
\frac{1}{2}
(P(\mathbf{f}^{v}_i \mid \mathbf{M}^{v}) \log \frac{P(\mathbf{f}^{v}_i \mid \mathbf{M}^{v})}{P(\mathbf{f}_{v}^i \mid \tilde{\mathbf{M}}^{v})} \\
+ 
P(\mathbf{f}^{v}_i \mid \tilde{\mathbf{M}}^{v}) \log  \frac{P(\mathbf{f}^{v}_i \mid \tilde{\mathbf{M}}^{v})}{P(\mathbf{f}^{v}_i \mid \mathbf{M}^{v})}),
\end{aligned}
\end{equation}
where $P\left(\mathbf{f} \mid \mathbf{M}\right)$ indicates a probability prediction of the instance $\mathbf{f}$ belonging to the cluster corresponding to each prototype in the memory bank $\mathbf{M}$. Note that, only predictions from a pair of memories with the same scale ($\{ \mathbf{M}^{v}, \tilde{\mathbf{M}}^{v} \}$ or $\{ \mathbf{M}^{r}, \tilde{\mathbf{M}}^{r} \}$) are used in Eq.~\ref{KLDivergencevva}. For a mini-batch, the consistency constraint loss 
$\mathcal{L}_{cc}$ is formulated as:
\begin{equation}\label{KLDivergenceSUM}
\begin{aligned}
    \mathcal{L}_{cc}=
    \mathcal{L}^{v}_{\mathbf{M}^v\leftrightarrow \tilde{\mathbf{M}}^v} + 
    \mathcal{L}^{\hat{v}}_{\mathbf{M}^v\leftrightarrow \tilde{\mathbf{M}}^v} +
    \mathcal{L}^{r}_{\mathbf{M}^v\leftrightarrow \tilde{\mathbf{M}}^v} \\
    + 
    \mathcal{L}^{v}_{\mathbf{M}^r\leftrightarrow \tilde{\mathbf{M}}^r} +
    \mathcal{L}^{\hat{v}}_{\mathbf{M}^r\leftrightarrow \tilde{\mathbf{M}}^r} + 
    \mathcal{L}^{r}_{\mathbf{M}^r\leftrightarrow \tilde{\mathbf{M}}^r},    
\end{aligned}
\end{equation}
where the superscript $p$ and  $a\leftrightarrow b$ in $\mathcal{L}^{p}_{\mathbf{M}\leftrightarrow \tilde{\mathbf{M}}}$ represents that, we constrain the predictions of feature $\mathbf{f}^p_i$ from the 
corresponding memory banks $\mathbf{M}$ and $\tilde{\mathbf{M}}$.

\subsection{Optimization}\label{loss function}

The total training loss $\mathcal{L}$ can be formulated as follows:
\begin{equation}\label{OverallLoss}
\mathcal{L}=
\mathcal{L}_{ms} +
\alpha * \mathcal{L}_{ma} +
\beta * \mathcal{L}_{cc},
\end{equation}
where $\mathcal{L}_{ms}$, $\mathcal{L}_{ma}$, $\mathcal{L}_{cc}$ are described in detail in Eq. \ref{Lms}, Eq. \ref{Lma} and Eq. \ref{KLDivergenceSUM}. $\alpha$ and $\beta$ are trade-off hyper-parameters to balance these three terms.

\begin{table*}
	\footnotesize
	\centering
   \caption{Comparison with the state-of-the-art VI-ReID methods on SYSU-MM01 dataset. It contains supervised and unsupervised ReID methods. Rank-k accuracy (\%), mAP(\%) and mINP(\%) are reported.}
	\label{table:SYSU-MM01}
	\setlength\tabcolsep{8pt}
	\vspace{0pt}
	\scalebox{1}[1]{
		\begin{tabular}{c|c|c|ccc|cc|ccc|cc}
			\toprule
			& \multicolumn{2}{c|}{SYSU-MM01 Settings} & \multicolumn{5}{c|}{All-search} & \multicolumn{5}{c}{Indoor-search} \\
			
			\cline{2-13}  &{Methods} &{Venue} & Rank-1 & Rank-10 & Rank-20 & mAP & mINP & Rank-1 & Rank-10 & Rank-20 & mAP & mINP\\
            \hline
            \multirow{10}{*}{\rotatebox{90}{Supervised}} & Zero-Pad\cite{wu2017rgb} & ICCV-17 & 14.80 & 54.12 & 71.33 & 15.95 & - & 20.58 & 68.38 & 85.79 & 26.92 & -  \\
			& eBDTR\cite{2020Bi} & TIFS-19 & 27.82 & 67.34 & 81.34 & 28.42 & - & 32.46 & 77.42 & 89.62 & 42.46 & - \\
			& AlignGAN\cite{wang2019rgb} & ICCV-19 & 42.40 & 85.0 & 93.7 & 40.70 & - & 45.90 & 87.60 & 94.40 & 54.30 & -\\
			& cm-SSFT\cite{lu2020cross} & TPAMI-20 & 47.70 & - & - & 54.10 & - & - & - & - & - & -\\
   			& AGW\cite{2021Deep} & CVPR-20 & 47.50 & 84.39 & 92.14 & 47.65 & 35.30 & 54.17 & 91.14 & 95.98 & 62.97 & 59.23\\
			& DDAG\cite{ye2020dynamic} & ECCV-20 & 54.75 & 90.39 & 95.81 & 53.02 & 39.62 & 61.02 & 94.06 & 98.41 & 67.98 & 62.61 \\
   			& CA\cite{2021channel} & ICCV-21 & 69.88 & 95.71 & 98.46 & 66.89 & 53.61 & 76.26 & 97.88 & 99.49 & 80.37 & 76.79 \\
            & MPANet\cite{wu2021discover} & CVPR-21 & 70.58 & 96.21 & 98.80 & 68.24 & - & 76.74  & 98.21 & 99.57 & 80.96 & -\\
            & MAUM\cite{2022MAUM} & CVPR-22 & 71.68 & - & - & 68.79 & - & 76.97 & - & - & 81.94 & - \\
            & CTFT\cite{2022CIFT} & ECCV-22 & \underline{74.08} & - & - & \underline{74.79} & - & \underline{81.82} & - & - & \underline{85.61} & - \\
            \hline
            \multirow{10}{*}{\rotatebox{90}{Unupervised}} 
            & SPCL \cite{2020spcl} & NIPS-20 & 18.37 & 54.08 & 69.02 & 19.39 & 10.99 & 26.83 & 68.31 & 83.24 & 36.42 & 33.05  \\
            & MMT \cite{2020MMT} & ICLR-20 & 21.47 & 59.65 & 73.29 & 21.53 & 11.50 & 22.79 & 63.18 & 79.04 & 31.50 & 27.66  \\
            & CAP \cite{2021CAP} & AAAI-21 & 16.82 & 47.60 & 61.42 & 15.71 & 7.02 & 24.57 & 57.93 & 72.74 & 30.74 & 26.15  \\
            & Cluster Contrast \cite{2021ClusterContrast} & arXiv-21 & 20.16 & 59.27 & 72.5 & 22.00 & 12.97 & 23.33 & 68.13 & 82.66 & 34.01 & 30.88  \\
            & ICE \cite{2021ICE} & ICCV-21 & 20.54 & 57.5 & 70.89 & 20.39 & 10.24 & 29.81 & 69.41 & 82.66 & 38.35 & 34.32  \\
            & PPLR \cite{2022PPLR} & CVPR-22 & 12.58 & 47.43 & 62.69 & 12.78 & 4.85 & 13.65 & 52.66 & 70.28 & 22.19 & 18.35\\
            & ISE \cite{2022ISE} & CVPR-22 & 20.01 & 57.45 & 72.50 & 18.93 & 8.54 & 14.22 & 58.33 & 75.32 & 24.62 & 21.74 \\
            & H2H \cite{2021H2H} & TIP-21 & 30.15 & 65.92 & 77.32 & 29.40 & - & - & - & - & - & -  \\   
            & OTLA \cite{2022OLTA} & ECCV-22 & 29.9 & - & - & 27.1 & - & 29.8 & - & - & 38.8 & - \\
            & ADCA \cite{2022ADCA} & MM-22 & \underline{45.51} & \underline{85.29} & \underline{93.16} & \underline{42.73} & \underline{28.29} & \underline{50.60} & \underline{89.66} & \textbf{96.15} & \underline{59.11} & \underline{55.17}  \\ 
            \hline
            & Ours & - & \textbf{53.14} & \textbf{89.61} & \textbf{96.74} & \textbf{48.16} & \textbf{32.41} & \textbf{55.21} & \textbf{91.44} & \underline{95.83} & \textbf{61.98} & \textbf{57.13}\\
            \bottomrule
		\end{tabular}}
\end{table*}

\begin{table*}
	\footnotesize
	\centering
      \caption{Comparison with the state-of-the-art VI-ReID methods on RegDB dataset. It contains supervised and unsupervised ReID methods. Rank-k accuracy (\%), mAP(\%) and mINP(\%) are reported.}
	\label{table:RegDB}
	\setlength\tabcolsep{8pt}
	\vspace{0pt}
	\scalebox{1}[1]{
		\begin{tabular}{c|c|c|ccc|cc|ccc|cc}
			\toprule
			& \multicolumn{2}{c|}{RegDB Settings} & \multicolumn{5}{c|}{Visible to Infrared} & \multicolumn{5}{c}{Infrared to Visible} \\
			
			\cline{2-13}  &{Methods} &{Venue} & Rank-1 & Rank-10 & Rank-20 & mAP & mINP & Rank-1 & Rank-10 & Rank-20 & mAP & mINP\\
            \hline
            \multirow{10}{*}{\rotatebox{90}{Supervised}} & Zero-Pad\cite{wu2017rgb} & ICCV-17 & 17.75 & 34.21 & 44.35 & 18.90 & - & 16.63 & 34.68 & 44.25 & 17.82 & -  \\
			& eBDTR\cite{2020Bi} & TIFS-19 & 34.62 & 58.96 & 68.72 & 33.46 & - & 34.21 & 58.74 & 68.64 & 32.49 & - \\
			& AlignGAN\cite{wang2019rgb} & ICCV-19 & 57.9 & - & - & 53.6 & - & 56.3 & - & - & 53.4 & -\\
			& cm-SSFT\cite{lu2020cross} & CVPR-20 & 72.3 & - & - & 72.9 & - & 71.0 & - & - & 71.7 & -\\
   			& AGW\cite{2021Deep} & TPAMI-21 & 70.05 & 86.21 & 91.15 & 66.37 & 50.19 & 70.49 & 87.21 & 91.84 & 65.90 & 51.24 \\
			& DDAG\cite{ye2020dynamic} & ECCV-20 & 69.34 & 86.19 & 91.49 & 63.46 & 49.24 & 68.06 & 85.15 & 90.31 & 61.80 & 48.62 \\
   			& CA\cite{2021channel} & ICCV-21 & 85.03 & 95.49 & 97.54 & 79.14 & 65.33 & 84.75 & 95.33 & 97.51 & 77.82 & 61.56 \\
            & MPANet\cite{wu2021discover} & CVPR-21 & 82.8 & - & - & 80.7 & - & 83.7 & - & - & 80.9 & -\\
            & MAUM\cite{2022MAUM} & CVPR-22 & 87.87 & - & - & 85.09 & - & 86.95 & - & - & 84.83 & - \\
            & CTFT\cite{2022CIFT} & ECCV-22 & \underline{91.96} & - & - & \underline{92.00} & - & \underline{90.30} & - & - & \underline{90.78} & - \\
            \hline
            \multirow{10}{*}{\rotatebox{90}{Unupervised}} 
            & SPCL \cite{2020spcl} & NIPS-20 & 13.59 & 26.98 & 34.88 & 14.86 & 10.36 & 11.70 & 25.53 & 32.82 & 13.56 & 10.09  \\
            & MMT \cite{2020MMT} & ICLR-20 & 25.68 & 42.23 & 54.03 & 26.51 & 19.56 & 24.42 & 41.21 & 51.89 & 25.59 & 18.66  \\
             & CAP \cite{2021CAP} & AAAI-21 & 9.71 & 19.27 & 25.6 & 11.56 & 8.74 & 10.21 & 19.91 & 26.38 & 11.34 & 7.92  \\
            & Cluster Contrast \cite{2021ClusterContrast} & arXiv-21 & 11.76 & 24.83 & 32.84 & 13.88 & 9.94 & 11.14 & 24.11 & 32.65 & 12.99 & 8.99  \\
            & ICE \cite{2021ICE} & ICCV-21 & 12.98 & 25.87 & 34.4 & 15.64 & 11.91 & 12.18 & 25.67 & 34.9 & 14.82 & 10.6  \\
            & PPLR \cite{2022PPLR} & CVPR-22 & 8.93 & 20.87 & 27.91 & 11.14 & 7.89 & 8.11 & 20.29 & 28.79 & 9.07 & 5.65  \\
            & ISE \cite{2022ISE} & CVPR-22 & 16.12 & 23.30 & 28.93 & 16.99 & 13.24 & 10.83 & 18.64 & 27.09 & 13.66 & 10.71  \\
            & H2H \cite{2021H2H} & TIP-21 & 23.81 & 45.31 & 54.00 & 18.87 & - & - & - & - & - & -  \\
            & OTLA \cite{2022OLTA} & ECCV-22 & 32.9 & - & - & 29.7 & - & 32.1 & - & - & 28.6 & - \\
            & ADCA \cite{2022ADCA} & MM-22 & \underline{67.20} & \underline{82.02} & \underline{87.44} & \underline{64.05} & \underline{52.67} & \underline{68.48} & \underline{83.21} & \underline{88.00} & \underline{63.81} & \underline{49.62}  \\ 
            \hline
            & Ours & - & \textbf{83.79} & \textbf{95.83} & \textbf{97.82} & \textbf{77.87} & \textbf{65.04} & \textbf{82.82} & \textbf{95.73} & \textbf{96.89} & \textbf{76.74} & \textbf{61.73} \\
            \bottomrule
		\end{tabular}}
\end{table*}

\section{Experiment}

\subsection{Dataset and Evaluation Protocol}

We evaluate our proposed method on two public VI-ReID datasets: SYSU-MM01~\cite{wu2017rgb} and RegDB~\cite{nguyen2017person}. All experiments follow the common evaluation protocols used for VI-ReID~\cite{wu2017rgb, ye2018hierarchical}. The Rank-k accuracy and mean Average Precision (mAP) are adopted as evaluation metrics. In addition, we report the mean Inverse Negative Penalty (mINP) metric proposed in~\cite{2021Deep}.

$ \bullet $ \textbf{SYSU-MM01} is a VI-ReID dataset, which includes 491 identities collected from four visible and two near-infrared cameras. The training set contains 395 identities with 22,258 RGB and 11,909 infrared images, while the test set contains 96 identities. Following~\cite{wu2017rgb}, we evaluate the proposed method under two search modes, the All-search mode and the Indoor-search mode. In both of these two modes, the query set contains all infrared testing images. In All-search mode, the gallery set contains all visible images, while in indoor-search mode, the gallery set only contains images captured from indoor cameras. 

$ \bullet $ \textbf{RegDB} dataset is captured by a pair of aligned visible and infrared cameras~\cite{nguyen2017person}. It contains 412 identities with 8240 images, where each identity has 10 infrared images and 10 visible images. We randomly select 206 identities for training and the remaining 206 identities for testing. We evaluate our method under the two testing modes on RegDB dataset: Visible-to-Infrared and Infrared-to-Visible, representing querying visible images from the infrared image gallery, and vice versa. 

\begin{table*}
	\footnotesize
	\centering
	\caption{Ablation study on individual components on SYSU-MM01 and RegDB datasets.}
	\label{table:AblationStudy}
	\setlength\tabcolsep{4pt}
	\vspace{0pt}
	\scalebox{1}[1]{
		\begin{tabular}{c|ccccc|ccc|ccc|ccc|ccc}
			\toprule
			\multirow{2}{*}{Index} & \multicolumn{5}{c|}{Components} & \multicolumn{3}{c|}{SYSU-MM01(All-search)} & \multicolumn{3}{c|}{SYSU-MM01(Indoor-search)} & \multicolumn{3}{c|}{RegDB(Visible to Infrared)} & \multicolumn{3}{c}{RegDB(Infrared to Visible)}\\

			\cline{2-18} & Baseline & BCCM & MBCCM & MSMA & CC & Rank-1 & Rank-10 & mAP & Rank-1 & Rank-10 & mAP & Rank-1 & Rank-10 & mAP & Rank-1 & Rank-10 & mAP\\
			\hline	
			1 & \checkmark & - & - & - & - & 35.02 & 75.31 & 33.93 & 38.50 & 79.26 & 46.35
            & 42.15 & 63.01 & 41.18 & 43.01 & 64.85 & 40.51 \\
			2 & \checkmark & \checkmark & - & \checkmark & - & 46.73 & 86.77 & 41.89 & 45.52 & 85.78 & 53.78
            & 68.20 & 86.07 & 64.65 & 69.66 & 88.11 & 64.24 \\
			3 & \checkmark & - & \checkmark & \checkmark & - & 51.12 & 88.88 & 46.33 & 52.76 & 89.76 & 59.93
            & 75.15 & 89.76 & 70.38 & 74.56 & 92.33 & 69.16\\
            \hline
			4 & \checkmark & - & \checkmark & \checkmark & \checkmark & \textbf{53.14} & \textbf{89.61} & \textbf{48.16} & \textbf{55.21} & \textbf{91.44} & \textbf{61.98}
            &  \textbf{83.79} & \textbf{95.83} & \textbf{77.87} & \textbf{82.82} & \textbf{95.73} & \textbf{76.74} \\

            \bottomrule
		\end{tabular}}
\end{table*}

\begin{figure*}
  \centering
  \includegraphics[width=18cm]{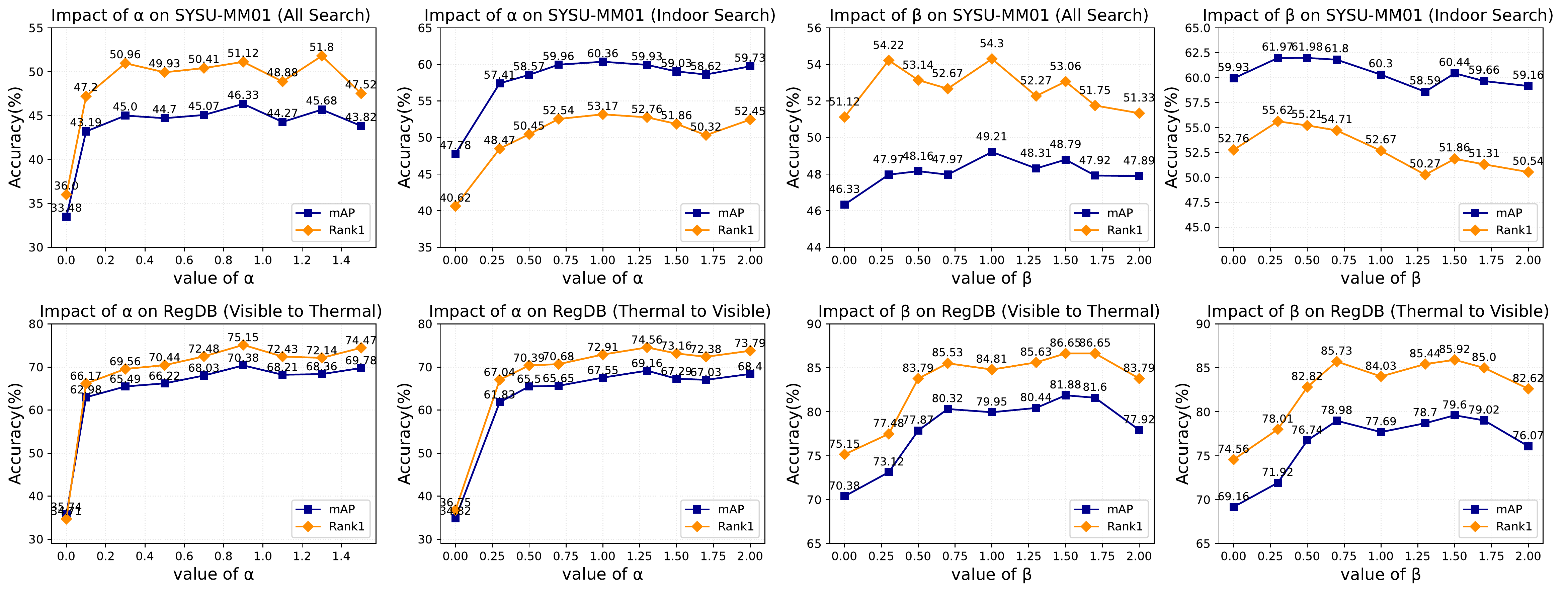}\\
  \caption{Performance of our framework with different values of $\alpha$ and $\beta$ on SYSU-MM01 and RegDB datasets.}\label{parameter-analysis}
\end{figure*}

\subsection{Implementation Details}

Our proposed method is implemented using PyTorch. We adopt ResNet50~\cite{2016resnet} pre-trained on ImageNet~\cite{2009imagenet} as the shared layers, while the settings of the two modality-specific shallow layers follow ADCA~\cite{2022ADCA}. After the shared layers, we add a GeM~\cite{2019GEM} pooling layer, followed by a batch normalization layer~\cite{2015batchnorm} and an L2-normalization layer, which will produce 2048-dimensional features. At the beginning of each epoch, we use DBSCAN~\cite{1996DBSCAN} for clustering to generate pseudo-labels.

The input image is resized to $288 \times 144$ for training. we perform random horizontal flipping, padding, random cropping, random erasing, linear transformation generator~\cite{2023LTG}, and channel argumentation~\cite{2021channel} for the training images.
Each mini-batch contains 12 person identities and 12 instances for each identity from each modality training set. We use the Adam optimizer for training the model with weight decay 5e-4. The initial learning rate is set to 3.5e-4 and decays 10 times at the 20th, 50th, and 70th epochs. In the previous 40 epochs, we pre-trained our model under two modalities independently. Our learning framework is executed after the 40th epoch. 
The temperature factor $\tau$ in Eq.\ref{Contrastive1} is set to 0.05 and the momentum updating factor $\mu$ in Eq.\ref{update} is set to 0.1 following Cluster-Contrast~\cite{2021ClusterContrast}. The trade-off hyper-parameters $\alpha$ and $\beta$ in Eq.\ref{OverallLoss} are set to 0.9 and 0.5.

\subsection{Comparison with State-of-the-art Methods}

We compare our proposed method with state-of-the-art supervised VI-ReID (SL-VI-ReID) and unsupervised VI-ReID (USL-VI-ReID) methods, as shown in Table \ref{table:SYSU-MM01} and Table \ref{table:RegDB}. We also report a subset of unsupervised single-modality ReID methods under the USL-VI-ReID setting, following ADCA \cite{2022ADCA}.\\
\textbf{Comparison with Unsupervised Methods.} Our method achieves 48.16\% mAP and 53.14\% Rank-1 on the SYSU-MM01 dataset (All search) and 77.87\% mAP and 83.79\% Rank-1 on the RegDB dataset (Visible to thermal), outperforming the state-of-the-art ADCA \cite{2022ADCA} by 5.43\% mAP on SYSU-MM01 and 13.82\% mAP on RegDB. Additionally, our method surpasses the most superior unsupervised single-modality Cluster Contrast \cite{2021ClusterContrast} by a large margin of 26.16\% mAP on SYSU-MM01 and 63.99\% mAP on RegDB, demonstrating the significance of cross-modality information interaction. In comparison with H2H \cite{2021H2H} and OTLA \cite{2022OLTA}, which train two encoders under different modalities independently in advance, our framework trains one cross-modality encoder throughout the entire process, which is more efficient and unified. Furthermore, our method does not use any extra dataset to pretrain the encoder, while H2H \cite{2021H2H} uses Market-1501 for pretraining.\\
\textbf{Comparison with Supervised Methods.} Our proposed method outperforms some supervised methods, including Zero-Pad \cite{wu2017rgb}, eBDTR \cite{2020Bi}, and AlignGAN \cite{wang2019rgb}, and is on par with the strong supervised ReID baseline AGW \cite{2021Deep} on SYSU-MM01 in Table \ref{table:SYSU-MM01}. From Table \ref{table:RegDB}, we can see that our method reaches an impressive performance on RegDB, exceeding a considerable number of supervised methods and approaching CA \cite{2021channel}. 

\subsection{Ablation Study}\label{ablation}

\begin{figure}
\centering
\includegraphics[width=0.47\textwidth]{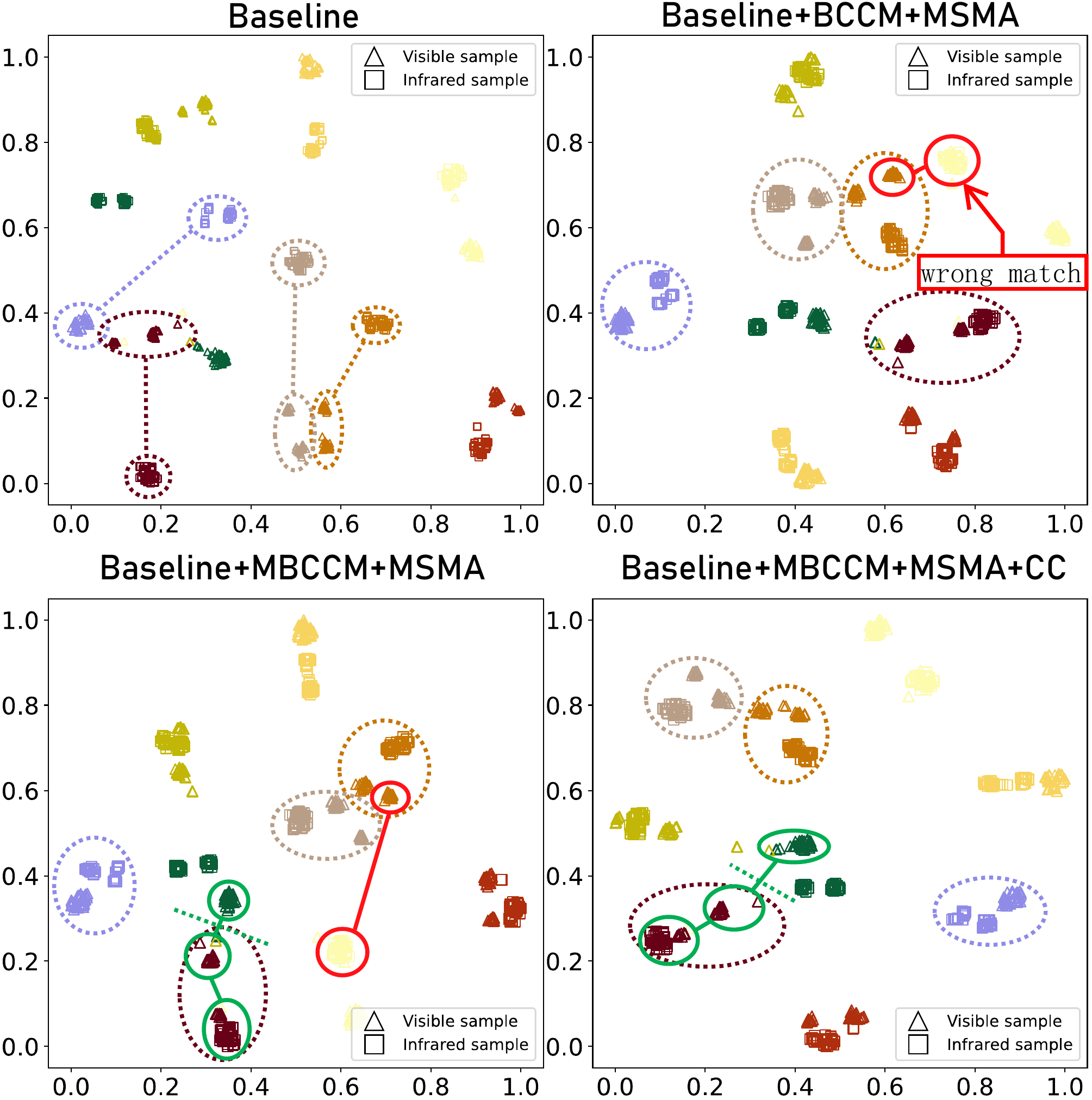}
\caption{The t-SNE\cite{2008tSNE} visualization of 10 randomly selected identities. Different colors represent different ground-truth identities. "$\mathbf{\square}$" denotes the samples from infrared modality while "$\mathbf{\triangle}$" denotes the samples from visible modality.} 
\vspace{-2mm}
\label{t-sne}
\end{figure}


In this subsection, we analyze the effectiveness of each component in our framework, including MBCCM in \ref{BCCM}, MSMA in \ref{MSMA}, and Consistency Constraint (CC) in \ref{CCP}. Our baseline is a Cluster-Contrast\cite{2021ClusterContrast} framework consisting of a dual-stream network as encoder and two modality-specific memory banks mentioned in section \ref{baseline}. Additionally, we compare the bilateral one-to-one cross-modality cluster matching (BCCM) introduced in Eq. \ref{optimize visible} in section \ref{BCCM} with our improved MBCCM method, to illustrate the benefits of the many-to-many matching paradigm. We conduct ablation studies on SYSU-MM01 and RegDB datasets, and the results are shown in Table \ref{table:AblationStudy}.\\
\textbf{Effectiveness of BCCM and MSMA.} The efficacy of BCCM and MSMA is revealed when comparing index 2 and index 1, which achieves +7.96\%/+11.71\% and +23.47\%/+26.05\% mAP/Rank-1 improvements on SYSU (All search) and RegDB (Visible to thermal) compared to the baseline. The BCCM provides a coarse one-to-one match to generate shared pseudo labels between heterogeneous data points, while MSMA promotes the encoder to generate modality-invariant features preliminarily under the effect of the partial correct match between relatively distinguishable clusters.\\
\textbf{Effectiveness of MBCCM.} Index 3 verifies the effectiveness of MBCCM, as it provides an evident boost of up to +4.44\%/+4.39\% and +5.73\%/+6.95\% in mAP/Rank-1 on the SYSU-MM01 and RegDB datasets, respectively, demonstrating the usefulness of such a many-to-many matching paradigm. An important factor contributing to performance improvement is that MBCCM enables each cluster to match multiple other clusters with high similarity, correcting some overconfident wrong matches in BCCM to a certain extent. \\
\textbf{Effectiveness of CC.} The CC module further improves performance by +1.83\%/+2.02\% and +7.49\%/+8.64\% in mAP/Rank-1 on the two datasets, respectively. It is noteworthy that the CC module obtains a considerable performance gain on the easier dataset RegDB. These results illustrate that the CC module significantly narrows the distance between cross-modal instances for a high-quality match while being sensitive to matching noise. 

\subsection{More Discussions}

\begin{figure}
\centering
\includegraphics[width=0.48\textwidth]{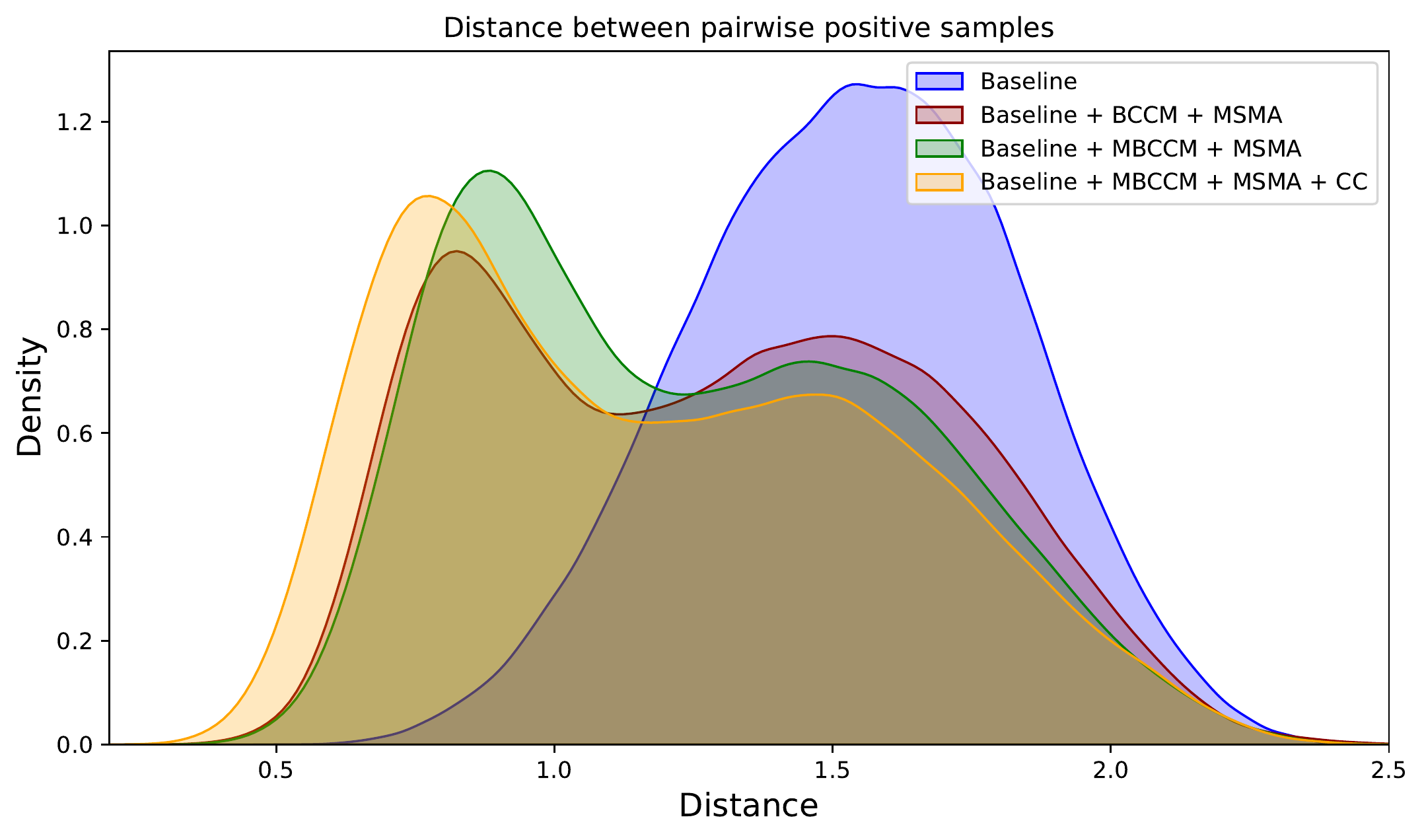}
 \vspace{-7mm}
\caption{The distribution of the distance between 200000 randomly selected pairwise positive samples in SYSU-MM01 dataset.  By integrating the proposed modules (i.e., BCCM, MSMA, MBCCM and CC ) into the training framework step-by-step, the peak of the distance distribution shifts to the left (shorter distance) gradually.
} 
\vspace{-3mm}
\label{analysis-3}
\end{figure}

\textbf{Parameter Analysis.} We analyze the impact of two key parameters in our method, i.e., the trade-off hyper-parameters $\alpha$ and $\beta$ in Eq.~\ref{OverallLoss}. We tune the value of each parameter while keeping the others fixed on SYSU-MM01 and RegDB. The results are shown in Figure \ref{parameter-analysis}. 
A too-large value of $\alpha$ leads to too much focus on modality-agnostic learning and the neglect of inter-class variation.
The upper in figure \ref{parameter-analysis} reveals that a too-large value of $\beta$ causes the performance to drop substantially in SYSU-MM01, which proves that the CC module is sensitive to matching noise.
Based on these experimental results, we set $\alpha = 0.9$ and $\beta = 0.5$ in our model.\\
\textbf{Visualization.} As shown in Figure \ref{t-sne}, we randomly select 10 identities from SYSU-MM01 dataset and plot the t-SNE\cite{2008tSNE} map, where the same identity has the same color. The BCCM and MSMA learning framework bring samples of the same identity from different modalities closer together, while there exist wrong matches marked with red circles. The MBCCM algorithm revises these wrong matches by connecting them to multi-clusters that probably contain the actual matched one. 
The consistency constraint module further shortens the distance of cross-modality clusters and improves the intra-class compactness.
Specifically, the green circles in the lower right in Figure \ref{t-sne} reveal that intra-class distances are shortened and inter-class distances are stretched under the cross-modality consistency constraint. \\
\textbf{Analysis of the Modality Gap.} We randomly selected 200,000 pairwise positive samples and computed the distribution of their Euclidean distance, as shown in Figure \ref{analysis-3}. 
By integrating the following modules (i.e., BCCM, MBCCM, MSMA, and CC) step-by-step into the training framework, the peak of the distance distribution shifts to the left  (shorter distance) gradually, illustrating the role of each part in our framework to reduce the modality gap.

\section{Conclusion}

In this paper, we propose a novel Bilateral Cluster Matching-based Learning framework for unsupervised visible-infrared person ReID. Our method connects cross-modality clusters using a Many-to-many Bilateral Cross-modality Cluster Matching algorithm through optimizing the maximum matching problem in a bipartite graph. We also propose a Modality-Specific and Modality-Agnostic learning framework, along with a cross-modality Consistency Constraint module, to reduce the modality gap between the two unlabeled sub-datasets. 
Finally, we hope that our study can help researchers to solve the USL-VI-ReID task from a new perspective.



\bibliographystyle{HLF}
\bibliography{HLF}

\end{document}